\documentclass[conference]{IEEEtran}
\IEEEoverridecommandlockouts
\usepackage{cite}
\usepackage{hyperref} 
\usepackage{amsmath,amssymb,amsfonts}
\usepackage{algorithmic}
\usepackage{graphicx}
\usepackage{textcomp}
\usepackage{xcolor}
\usepackage{empheq}
\usepackage{tikz}
\usepackage{booktabs} 
\usepackage{tabularx} 
\usepackage{amsmath}
\usepackage{tcolorbox} 
\usepackage{xcolor} 
\usepackage{amsmath}
\usepackage{amsfonts}

\hypersetup{
    colorlinks=false, 
    linkcolor=green, 
    citecolor=green, 
    urlcolor=green,  
    pdfborder={0 0 0} 
}

\def\BibTeX{{\rm B\kern-.05em{\sc i\kern-.025em b}\kern-.08em
    T\kern-.1667em\lower.7ex\hbox{E}\kern-.125emX}}
\begin{document}

\title{DAAL: Density-Aware Adaptive Line Margin Loss for Multi-Modal Deep Metric Learning\\
\thanks{Identify applicable funding agency here. If none, delete this.}
}

\author{
\IEEEauthorblockN{1\textsuperscript{st} Hadush Hailu Gebrerufael}
\IEEEauthorblockA{\textit{Computer Science} \\
\textit{MIU}\\
Iowa, USA \\
hadush.gebrerufael@miu.edu}
\and

\IEEEauthorblockN{2\textsuperscript{nd} Anil Kumar Tiwari}
\IEEEauthorblockA{\textit{Science and Technology} \\
\textit{Pokhara University}\\
Pokhara, Nepal \\
anil@student.pu.edu.np}
\and

\IEEEauthorblockN{3\textsuperscript{rd} Gaurav Neupane}
\IEEEauthorblockA{\textit{Computer Science} \\
\textit{MIU}\\
Iowa, USA \\
gaurav.neupane@miu.edu}
\and

\IEEEauthorblockN{4\textsuperscript{th} Goitom Ybrah Hailu}
\IEEEauthorblockA{\textit{Eng. for Innovative Medicine} \\
\textit{Universita di Verona}\\
Verona, Italy \\
goitomybrah.hailu@studenti.univr.it}

}

\maketitle

\begin{abstract}
Multi-modal deep metric learning is crucial for effectively capturing diverse representations in tasks such as face verification, fine-grained object recognition, and product search. Traditional approaches to metric learning, whether based on distance or margin metrics, primarily emphasize class separation, often overlooking the intra-class distribution essential for multi-modal feature learning. In this context, we propose a novel loss function called Density-Aware Adaptive Margin Loss(DAAL), which preserves the density distribution of embeddings while encouraging the formation of adaptive sub-clusters within each class. By employing an adaptive line strategy, DAAL not only enhances intra-class variance but also ensures robust inter-class separation, facilitating effective multi-modal representation. Comprehensive experiments on benchmark fine-grained datasets demonstrate the superior performance of DAAL, underscoring its potential in advancing retrieval applications and multi-modal deep metric learning.
\end{abstract}


\section{Introduction}
Metric learning which focus on learning discriminative feature in high-dimensional space has been extensively studied in the past decades due to its broad range of applications, e.g., k-nearest neighbor classification \cite{b7}, image retrieval \cite{b8} and clustering \cite{b21}. The main aim of metric learning is to create an effective distance metric that groups samples of the same class more closely together, while increasing the separation between samples of different classes. Numerous algorithms have been introduced to achieve this \cite{b6, b12, b32, b7}.

Early work in DML, referred to as classical DML methods, relied on hand-crafted features to represent examples, with the primary objective of learning a feature mapping that projects these examples from the original feature space to a new, more discriminative space. While these methods were effective at deriving useful metrics from the input features, the use of hand-crafted features often led to information loss, and the high dimensionality of the input space posed significant challenges. Although approaches like PCA were employed to reduce the dimensionality \cite{b7}, and later strategies were introduced to lower computational costs \cite{b12,b31}, these issues persisted.

With the rise of Deep Neural Networks (DNNs) and Deep Convolutional Neural Networks (CNNs), researchers began focusing on learning data embeddings directly through neural networks \cite{b6, b32}, eliminating the need for manual feature extraction. This method, known as Deep Metric Learning(DML), has resulted in substantial performance improvements \cite{b32}, opening a new frontier in the field. By leveraging the ability to learn non-linear feature representations, deep metric learning has demonstrated outstanding results across a wide range of tasks, including highlight detection \cite{b30, b66}, zero-shot image classification \cite{b63, b18}, clustering \cite{b3}, image retrieval \cite{b13, b50, b19, b49, b54}, visual product search \cite{b10, b56, b8, b55, b33, b34}, face recognition \cite{b25, b17, b47, b5, b64, b42, b28}, person re-identification \cite{b57, b4}, feature matching \cite{b46}, fine-grained image classification \cite{b16, b45}, zero-shot learning \cite{b35,b43}, and collaborative filtering \cite{b36}. Moreover, deep metric learning has been successfully applied to 3D retrieval tasks \cite{b23}.

Loss functions are essential to deep metric learning. While the traditional softmax loss is widely utilized, it often falls short in generating highly discriminative feature vectors \cite{b9, b39, b17}. This limitation prompted researchers to develop improved loss functions that enhance intra-class compactness and inter-class separability, ultimately boosting recognition accuracy and generalization. These efforts led to the emergence of distance-based DML loss functions, particularly contrastive loss (also known as the Siamese network) \cite{b25, b59, b29, b65}. Although contrastive loss has demonstrated significant success, it necessitates precise real-valued pairwise similarities or distances in the training data, which are frequently unavailable in real-world applications. To address this issue, triplet loss \cite{b9, b2, b7} was introduced by Weinberger and Saul. This loss function encourages the features of data points with the same identity to be closer together than those with different identities by leveraging relative similarity among pairs. Triplet loss has since been widely adopted in various tasks, including image retrieval \cite{b9, b38} and face recognition \cite{b5}. Furthermore, contributions such as N-pair loss \cite{b19} and various refinements of triplet loss \cite{b17, b60, b5} have further enhanced the learning of robust feature representations.

However, triplet loss introduces a challenge due to its requirement for triplet constraints, which can scale up to $O(n3)$, where n represents the number of original examples—making it computationally infeasible for large-scale datasets. And also since most deep models are trained using Stochastic Gradient Descent (SGD), which processes only mini-batches of data per iteration, the triplet loss strategy struggles to capture the full data distribution. The limited size of mini-batches constrains the available information compared to the entire dataset.

To address this, algorithms need to devise effective sampling strategies to construct mini-batches and sample triplet constraints from them. A straightforward approach is to increase the mini-batch size \cite{b5}, but this comes with challenges, such as GPU memory limitations and increased complexity in sampling triplets. An alternative method proposed by \cite{b11} involves generating mini-batches from neighboring classes. Additionally, various sampling strategies have been developed to improve constraint selection: \cite{b32} suggests sampling semi-hard negative examples, \cite{b8} uses all negative examples within a margin for each positive pair, \cite{b1} introduces distance-weighted sampling based on proximity to the anchor example, and \cite{b52} selects hard triplets with a dynamic violation margin from a hierarchical class-level tree. Despite these improvements, all these strategies may still fail to accurately capture the distribution of the entire dataset.

The current state-of-the-art in deep metric learning predominantly relies on margin-based softmax loss functions, commonly referred to as margin-based loss functions. These approaches enhance feature discrimination by introducing a margin to each identity, thereby increasing class separation. Prominent methods include SphereFace \cite{b39}, which applies a multiplicative angular margin, CosFace \cite{b17}, which introduces an additive cosine margin, and ArcFace \cite{b9, b25}, which incorporates an additive angular margin to further boost class separation. These techniques assume that all classes are adequately represented with sufficient samples to accurately describe their distributions, thus employing a constant margin for all classes is deemed sufficient.

To address the issue of distribution, strategies have been proposed to enlarge intra-class variation while penalizing the overlap of representation distributions across different classes in the embedding space. This approach aims to ensure that the embedding representations are sufficiently spread out to fully leverage the expressive power of the embedding space. Density adaptability \cite{b20} focuses on preserving inter-class density by maintaining the relationships between class densities in both the original and embedded spaces, though this introduces additional computational complexity. SoftTriple Loss \cite{b44}, which maintains multiple centers, was also proposed to improve this process. However, the former method is not applicable to datasets without prior information, and the latter method may inadvertently force points to cluster closely to sub-clusters while ignoring continuous distribution.

As a result of the aforementioned challenges, our study proposes a novel loss function designed to act as a regularizer capable of capturing continuous distributions without forcing points into local clusters while maintaining inter-class separability. This approach not only ensures clear separation between classes but also preserves the inherent distribution of each cluster. We achieve this by implementing an "elastic line" mechanism, which adapts to the class distribution by mapping it to an elastic line based on the variance of the class, allowing for flexible representation of the data structure.

The contributions of this paper are as follows:
\begin{itemize} \item Introduction and investigation of major loss functions in deep metric learning, explaining how they differ from the proposed DAAL (Density-Aware Elastic Line margin loss). \item Proposal of a novel loss function termed DAAL, a density-aware elastic line margin loss, demonstrating superior state-of-the-art results compared to other alternatives. \item Demonstration of the ease of implementing DAAL in pretrained Convolutional Neural Networks (CNNs), enabling straightforward integration into existing architectures. \end{itemize}

The remainder of this paper is organized as follows: Section II reviews related works on various types of deep metric learning. Section III introduces our novel regularizer loss function, along with an analysis and comparison to similar loss functions. Section IV presents performance comparisons on benchmark datasets. Finally, Section V concludes the paper and discusses potential future research directions.

\section{Related works}
In this section, we analyze state-of-the-art loss functions relevant to deep metric learning, highlighting both their strengths and limitations. While an exhaustive review of previous work is beyond the scope of this paper, we refer readers to an in-depth survey on metric learning \cite{b48} for further details. Traditionally, deep metric learning has been driven by two main categories: distance-based loss functions and margin-based loss functions. We will discuss recent advancements in these areas. Additionally, we introduce a third emerging category, known as distribution-aware loss functions, which, although not extensively covered in the literature, will be explored in this work.

\subsection{Distance-based Loss functions}
Historically, metric-based methodologies have been prevalent in image analysis. These techniques aim to establish a similarity metric for a set of images through deep metric learning networks \cite{b46}. Their primary objective is to position visually similar images closer in an embedding space while ensuring that visually dissimilar images are separated. Two principal categories of metric-based losses have been developed for different contexts: contrastive loss and triplet loss. The contrastive loss \cite{b22} facilitates network training by predicting whether pairs of samples belong to the same class, achieved by minimizing the embedding distance for positive pairs and maximizing it for negative pairs. This approach is frequently described as distance metric learning.

Conversely, triplet loss operates on triplet samples composed of an anchor, a positive, and a negative sample. This framework was initially applied in the FaceNet architecture \cite{b51}, where the goal is to minimize the distance between the anchor and the positive sample while maximizing the distance between the anchor and the negative sample.

Despite their theoretical merits, employing contrastive and triplet losses in deep representation learning can result in slow convergence and instability, as the optimization process only considers one negative sample per update \cite{b5, b16, b40}. To mitigate these challenges, Sohn et al. \cite{b16} proposed the (N + 1)-tuplet loss, which enhances training convergence by selecting a positive sample from a set of negative samples, thereby incorporating the distances between the anchor and multiple negatives. However, this method results in a quadratic increase in the sample count per batch, leading to substantial expansion in the sample space. To address these limitations, Wen et al. \cite{b40} introduced center loss, which concurrently learns the centroid for each class and penalizes the distances between feature representations and their respective class centers. Despite these advancements, center loss and analogous strategies exhibit limited efficacy in addressing the open set problem in face recognition tasks \cite{b5}.

\subsection{Margin-based Loss functions}
In recent years, a variety of margin-based softmax loss functions have been introduced, aiming to enhance the discriminative ability of features in deep learning models, especially for tasks like face recognition. These loss functions, including several widely recognized methods, incorporate additional constraints in the angular space to create a more discriminative decision boundary than traditional deep metric learning approaches \cite{b37, b32, b35, b36, b61, b60, b41, b67, b43}.

One of the early breakthroughs in this area was proposed by Liu et al. \cite{b60} with the introduction of large-margin softmax loss (L-softmax). This method leverages a piecewise function to maintain the monotonicity of the cosine function, directly fostering both intra-class compactness and inter-class separation. Building on this, Liu et al. \cite{b41} developed SphereFace, which normalizes the weight matrix in the final fully connected layer to geometrically constrain the features on a hypersphere manifold. While this method offers strong inter-class separability, its reliance on a multiplicative angular penalty margin made it difficult to optimize during training.

To mitigate the training challenges posed by SphereFace, CosFace \cite{b43} was proposed, reframing the classification problem using cosine distance. By introducing an additive cosine margin between the deep features and class weights, CosFace simplifies the learning process. However, the challenge lies in selecting an optimal cosine margin that balances intra-class compactness with inter-class separability. Following this, Deng et al. \cite{b32} introduced ArcFace, which added a more intuitive geometric interpretation of the margin by applying an additive angular margin, significantly boosting performance.

He et al. \cite{b68} highlighted a critical issue in traditional softmax approaches, pointing out that intra-class and inter-class objectives are inherently entangled. Optimizing one objective can inadvertently relax the other, creating a challenge for fine-tuning models to achieve both compactness within classes and separation between them. To address this, several adaptive methods were introduced. For instance, AdaptiveFace \cite{b61} and Dyn-arcFace \cite{b36} focus on class imbalances by learning adaptive margins for each class, providing better handling of underrepresented classes. CurricularFace \cite{b35}, on the other hand, uses curriculum learning to gradually adjust the balance between easy and hard examples during training, further refining the learning process. Additionally, KappaFace \cite{b67} offers a unique approach by modulating the positive margins based on class imbalance and difficulty level.

Among these innovations, AdaFace \cite{b69} introduced the notion of quality-aware training, where image quality is taken into account by adjusting the margin adaptively based on the feature norms. This approach is particularly effective in handling varying image qualities, which is crucial in real-world face recognition scenarios.
In conclusion, margin-based softmax loss functions have made remarkable strides in enhancing the performance of deep face recognition systems. By effectively minimizing intra-class variability and increasing inter-class separability, these methods have become integral to improving the accuracy and robustness of modern face recognition models. As a result, they are now more commonly adopted in practical face recognition applications, marking a significant advancement in the field.

\subsection{Distribution-based Loss functions}
Traditional margin-based and distance-based loss functions focus on enhancing inter-class separation while minimizing intra-class variance. In contrast, distribution-based loss functions aim to optimize inter-class separation while maximizing intra-class variations. Approaches like density adaptability \cite{b20} seek to increase intra-class variation and reduce the overlap of representation distributions across classes in the embedding space, ensuring that embeddings are well-distributed. SoftTriple Loss \cite{b44} also aims to improve class representation by maintaining multiple centers. However, these methods face limitations; density adaptability is unsuitable for datasets lacking prior information, and SoftTriple Loss may inadvertently lead to tight clustering within sub-clusters, neglecting continuous distribution.

Our study presents a novel loss function that acts as a regularizer, capturing continuous distributions without enforcing local clustering while maintaining inter-class separability. We implement an "elastic line" mechanism that adapts to class distributions by mapping them to an elastic line based on class variance, allowing flexible data representation. Unlike existing adaptive methods, our approach does not require prior information, and it retains the original data distribution while embedding it in high-dimensional space. Furthermore, our method can be easily integrated into existing architectures, such as softmax, enhancing overall model performance.

\section{Methodology}

In this section, we will first examine the mathematical foundations and intricacies of the common loss functions and regularizers employed in modern state-of-the-art Deep Metric Learning (DML). We will then provide a detailed introduction to our proposed loss function, followed by an analysis of the impact of hyperparameters on performance and strategies for fine-tuning them for optimal results.

\subsection{Review on Loss Functions and Regularizers}

To ensure consistency in the definitions of various loss functions and regularizers discussed, we establish the following terms that are commonly used throughout the section:

\begin{itemize}
    \item $N$: Sample size (number of examples/samples in the dataset).
    \item $C$: Number of classes/labels.
    \item $d$: Dimension of the embedding space/feature representation.
    \item $\mathbf{x}_i$: The $i$-th sample (embedding or input feature vector).
    \item $y_i$: The true label of the $i$-th sample.
    \item $\mathbf{W} \in \mathbb{R}^{d \times C}$: Weight matrix of the last fully connected layer, where each column $\mathbf{w}_j$ represents the weight vector for class $j$.
    \item $\mathbf{w}_{y_i}$: Weight vector corresponding to class $y_i$.
    \item $\theta_{y_i}$: Angle between the embedding $\mathbf{x}_i$ and the weight vector $\mathbf{w}_{y_i}$.
    \item $s$: Scaling factor used in normalization.
    \item $m$: Margin used in margin-based loss functions.
    \item $D(\cdot)$: Distance function, typically squared Euclidean distance/l2 norm.
\end{itemize}

\subsubsection{SoftMax}

The SoftMax function is pivotal in translating raw model outputs (logits) into probabilities for multi-class classification tasks. Given an input feature embedding $\mathbf{x}_i$ for the $i$-th sample, alongside its true label $y_i$, the SoftMax operation computes the likelihood of correctly predicting $y_i$.

It does so by exponentiating the dot product between the embedding $\mathbf{x}_i$ and the corresponding weight vector $\mathbf{w}_{y_i}$ for the true class $y_i$. This exponentiated term,
\[
\exp(\mathbf{w}_{y_i}^\top \mathbf{x}_i),
\]
quantifies the compatibility between the input $\mathbf{x}_i$ and the class label $y_i$. To create a valid probability distribution, this value is divided by the sum of the exponentiated dot products across all class weight vectors:
\[
\sum_{j=1}^{C} \exp(\mathbf{w}_j^\top \mathbf{x}_i),
\]
where $[\mathbf{w}_1, \dots, \mathbf{w}_C] \in \mathbb{R}^{d \times C}$ denotes the weight matrix from the final fully connected layer, with $C$ being the number of classes, $d$ the dimensionality of the embeddings, and $\mathbf{w}_j$ the weight vector for class $j$. This results in the following probability expression:
\[
P_i \Leftrightarrow P(Y = y_i \mid \mathbf{x}_i) = \frac{\exp(\mathbf{w}_{y_i}^\top \mathbf{x}_i)}{\sum_{j=1}^{C} \exp(\mathbf{w}_j^\top \mathbf{x}_i)}.
\]

For a batch of size $N$, the SoftMax loss function is defined as:
\begin{flushleft} 
\fbox{%
    \begin{minipage}{0.5\textwidth} 
        \vspace{10pt} 
         \begin{equation} 
        \zeta_{s} = -\frac{1}{N} \sum_{i=1}^{N} \log{P_i} \Leftrightarrow -\frac{1}{N} \sum_{i=1}^{N} \log \frac{\exp(\mathbf{w}_{y_i}^\top \mathbf{x}_i + b_{y_i})}{\sum_{j=1}^{C} \exp(\mathbf{w}_j^\top \mathbf{x}_i + b_j)}.
        \end{equation}
        \vspace{10pt} 
    \end{minipage}%
}
\end{flushleft}

\subsubsection{Normalized Softmax Loss}

The normalized Softmax loss transoforms $\mathbf{w}_j^\top \mathbf{x}_i + b_j$ to $\|\mathbf{w}_j\| \|\mathbf{x}_i\| \cos \theta_j,$ and ensures that the weight vectors \(\mathbf{w}_j\) and the input features \(\mathbf{x}_i\) reside on the unit hypersphere. This is accomplished by performing \(L_2\) normalization on both the weight vectors and the features, followed by the introduction of a scaling factor \(s\) on the input features.

The normalized Softmax loss can be represented as:
\begin{flushleft} 
\fbox{%
    \begin{minipage}{0.5\textwidth} 
        \vspace{10pt} 
        \begin{equation}
            \zeta_{\text{ns}} = - \frac{1}{N} \sum_{i=1}^{N} \log \left( \frac{\exp(s \cdot \cos(\theta_{y_i}))}{\sum_{j=1}^{C} \exp(s \cdot \cos(\theta_j))} \right)
        \end{equation}
        \vspace{10pt} 
    \end{minipage}%
}
\end{flushleft}

In this equation, \(\theta_{y_i}\) represents the angle between the embedding \(\mathbf{x}_i\) and its corresponding weight vector \(\mathbf{w}_{y_i}\).

\subsubsection{Margin-Based Softmax Losses (SphereFace, CosFace, ArcFace)}
To enhance the discriminative power of the learning embeddings various margin-based Softmax losses introduce a margin to increase the inter-class variance and reduce intra-class variance. This is achieved by introducing a margin 
between the target score and non-target scores. 

The general form of margin-based loss functions can be formulated as follows:
\begin{flushleft} 
\fbox{%
    \begin{minipage}{0.5\textwidth} 
        \vspace{10pt} 
        \begin{equation}
            \zeta_{m} = \frac{1}{N} \sum_{i=1}^{N} \log \left( \frac{e^{s \cdot g(m, \theta_{y_i})}}{e^{s \cdot g(m, \theta_{y_i})} + \sum_{j=1, j \neq y_i}^{n} e^{s \cdot \cos \theta_j}} \right)
            \label{eq:margin}
        \end{equation}
        \vspace{10pt} 
    \end{minipage}%
}
\end{flushleft}

where $g(m, \theta_{y_i})$ is the introduced margin function. For instance: 
\begin{itemize}
    \item \textbf{SphereFace}: The margin function is defined as \( g(m_1, \theta_y) = \cos(m_1 \theta_y) \), where \( m_1 \) is a multiplicative angular margin, satisfying \( m_1 \geq 1 \) and is an integer.
    
    \item \textbf{CosFace}: The margin function is expressed as \( g(m_2, \theta_y) = \cos(\theta_y) - m_2 \), where \( m_2 \) is an additive cosine margin, with \( m_2 \geq 0 \).
    
    \item \textbf{ArcFace}: The margin function is formulated as \( g(m_3, \theta_y) = \cos(\theta_y + m_3) \), where \( m_3 \) is an additive angular margin, and \( m_3 \geq 0 \).
\end{itemize}

\subsubsection{Triplet Loss}
Triplet loss, as indicated by its name, is computed over a triplet of training samples $(\mathbf{x}_a^i, \mathbf{x}_+^i, \mathbf{x}_-^i)$, where the samples $(\mathbf{x}_a^i, \mathbf{x}_+^i)$ belong to the same class, and $(\mathbf{x}_a^i, \mathbf{x}_-^i)$ belong to different classes. The sample $\mathbf{x}_a^i$ serves as the anchor for the triplet. Intuitively, the triplet loss seeks to create an embedding space where the distance between samples of the same class (i.e., $\mathbf{x}_a^i$ and $\mathbf{x}_+^i$) is smaller than the distance between samples from different classes (i.e., $\mathbf{x}_a^i$ and $\mathbf{x}_-^i$) by at least a margin $m$. The triplet loss is formally expressed as:

\begin{flushleft}
    \setlength{\fboxrule}{1pt} 
    \setlength{\fboxsep}{10pt} 
    \fbox{%
        \begin{minipage}{0.45\textwidth} 
            \vspace{10pt} 
            \begin{equation}
                \begin{split}
                    \zeta_{t} &= \sum_{i=1}^{N} \max\left(0, m + D\left(f(\mathbf{x}_a^i), f(\mathbf{x}_+^i)\right) \right. \\
                    &\quad - \left. D\left(f(\mathbf{x}_a^i), f(\mathbf{x}_-^i)\right)\right)
                \end{split}
                \label{eq:triplet_loss}
            \end{equation}
            \vspace{10pt} 
        \end{minipage}
    }
\end{flushleft}

where $f(\cdot)$ represents the feature embedding generated by the neural network, $D(\cdot)$ is the distance function (such as squared Euclidean distance), $N$ is the number of triplets in the dataset, and $i$ represents the $i$-th triplet.

\subsubsection{Center Loss}
Center loss is used to minimize the intra-class variance by learning a center $\mathbf{c}_{y_i}$ for each class $y_i$ and penalizing the distance between the features $\mathbf{x}_i$ and their corresponding class center $\mathbf{c}_{y_i}$. The center loss is formulated as:

\begin{flushleft} 
\fbox{%
    \begin{minipage}{0.47\textwidth} 
        \vspace{5pt} 
        \begin{equation}
            \zeta_{c} = \frac{1}{2} \sum_{i=1}^{N} D(\mathbf{x}_i, \mathbf{c}_{y_i})
        \end{equation}
        \vspace{10pt} 
    \end{minipage}%
}
\end{flushleft}

where $D(\cdot)$ is the Euclidean distance between the embedding $\mathbf{x}_i$ and the center $\mathbf{c}_{y_i}$.

\subsubsection{Triplet-Center Loss (TCL)}
The primary aim of Triplet-Center Loss (TCL) is to combine the strengths of triplet loss and center loss, effectively minimizing intra-class variability while maximizing inter-class separation in the feature space. Given a training dataset represented as $\{(\mathbf{x}_i, y_i)\}_{i=1}^{N}$, each sample $\mathbf{x}_i \in X$ is paired with a label $y_i \in \{1, 2, \dots, |Y|\}$, and is embedded into a $d$-dimensional space using the neural network function $f_\theta(\cdot)$. In the TCL framework, all features from the same class are associated with a shared center, leading to the set of centers defined as $C = \{\mathbf{c}_1, \mathbf{c}_2, \dots, \mathbf{c}_{|Y|}\}$, where each center $\mathbf{c}_y \in \mathbb{R}^d$ corresponds to class $y$ and the centers are updated during each training iteration using a mini-batch of size $N$.

\begin{flushleft}
    \setlength{\fboxrule}{1pt} 
    \setlength{\fboxsep}{5pt} 
    \fbox{%
        \begin{minipage}{0.47\textwidth} 
            \vspace{10pt} 
            \begin{equation}
                \zeta_{tcl} = \sum_{i=1}^{N} \max \left( D(f_i, \mathbf{c}_{y_i}) + m - \min_{j \neq y_i} D(f_i, \mathbf{c}_j), 0 \right)
                \label{eq:TCL}
            \end{equation}
            \vspace{10pt} 
        \end{minipage}
    }
\end{flushleft}

Here, $D(\cdot)$ represents the squared Euclidean distance:

\[
    D(f_i, \mathbf{c}_{y_i}) = \frac{1}{2} \|\mathbf{f}_i - \mathbf{c}_{y_i}\|_2^2
    \label{eq:euclidean_distance}
\]

TCL operates by ensuring that the distance from a sample to its corresponding class center $\mathbf{c}_{y_i}$ is minimized relative to the distances to the centers of other classes by a margin $m$.

\section{Density-Aware Adaptive Line Loss (DAAL) for Deep Metric Learning}
In this work, we introduce the Density-Aware Adaptive Line (DAAL) loss, which seeks to minimize intra-class variance and maximize inter-class separability in deep metric learning problems. The DAAL loss leverages class-specific line segments to dynamically adapt to the density of data points within each class. These line segments, denoted as $\overrightarrow{AB}$, are formed between two vertices, ${A}$ and ${B}$, for each class, with vertex positions dynamically adjusted based on class density and embedding variance.

\subsection{Initialization}

We define a line segment $\overrightarrow{AB}$ for each class, where:
- ${A} \in \mathbb{R}^{d}$ is the inner vertex of the line segment, initialized randomly in the embedding space.
- ${B} \in \mathbb{R}^{d}$ is the outer vertex of the line segment, initialized to lie on a unit vector $\hat{v}$, extending a fixed length $L$ from ${A}$. The initial direction $\hat{v}$ is sampled from a normal distribution and normalized as follows:

\[
\hat{v} = \frac{{v}}{\|{v}\|_2}
\]
where ${v}$ is a randomly sampled direction vector.

The vertex ${B}$ is then placed at a distance $L$ from ${A}$:
\[
{B} = {A} + L \cdot \hat{v}
\]

\subsection{Class Centroids and Variance}
We first compute the class centroids ${c}_k$ and variances $\sigma_k$ for each class $k \in \{1, \ldots, C\}$ as:
\[
{c}_k = \frac{1}{N_k} \sum_{y_i = k} {e}_i
\]
\[
\sigma_k^2 = \frac{1}{N_k} \sum_{y_i = k} \|{e}_i - {c}_k\|^2
\]
where $N_k$ is the number of samples in class $k$, ${e}_i \in \mathbb{R}^{d}$ of the training samples and their respective class labels $y_i \in \{1, \ldots, C\}$.

\subsection{Adaptive Line Construction}

\begin{figure*}[htpb]
    \centering
    \setlength{\fboxrule}{0.5pt} 
    \setlength{\fboxsep}{5pt} 
    \fbox{\includegraphics[width=\textwidth]{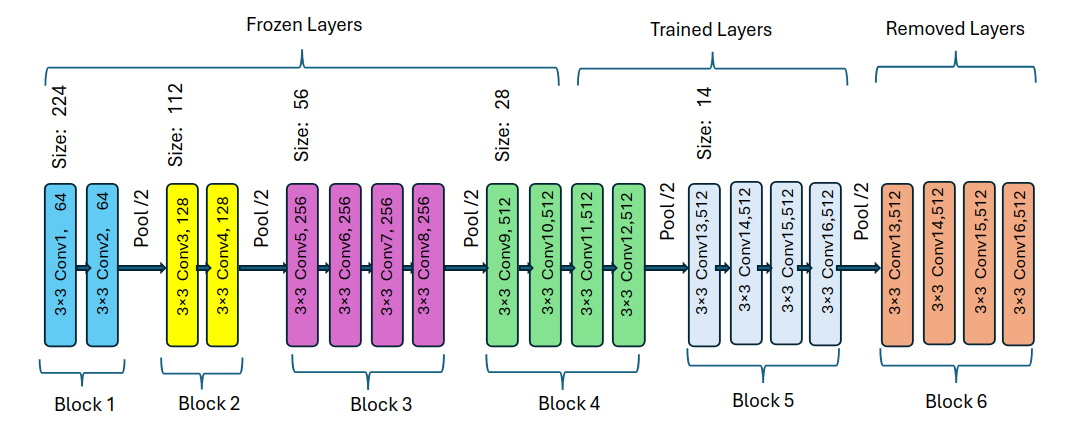}} 
    \caption{VGG-Net/VGG-19 Model Architecture}
\end{figure*}

The position of vertices ${A}_k$ and $ {B}_k$ for each class $k$ is then adjusted based on the class centroids and the class variance. Specifically:
\[
 {A}_k =  {c}_k - \eta \cdot \sigma_k
\]
\[
 {B}_k =  {c}_k + \eta \cdot \sigma_k
\]
where $\eta$ is a variance factor that controls the degree to which the line segment adapts to the intra-class variance. The vertex $ {A}_k$ is positioned within the dense region of the class, while $ {B}_k$ is placed outside the dense region, effectively allowing the line segment $\overrightarrow{AB}$ to expand or shrink based on the variance of the class.

\subsection{Distance Calculations}

For each embedding $ {e}_i$, we calculate the Euclidean distance to both vertices $ {A}_{y_i}$ and $ {B}_{y_i}$ corresponding to the correct class $y_i$:
\[
d( {e}_i,  {A}_{y_i}) = \| {e}_i -  {A}_{y_i}\|
\]
\[
d( {e}_i,  {B}_{y_i}) = \| {e}_i -  {B}_{y_i}\|
\]
We then calculate the distances from the embedding to the closest point on the line segment $\overrightarrow{AB}$ for each class, and for every incorrect class, we compute the minimum distance.

\subsection{Intra-Class and Inter-Class Losses}

The intra-class loss is defined as the squared distance between each embedding and the closest point on the line segment $\overrightarrow{AB}$ of the correct class:
\[
\text{Loss}_{\text{intra}} = \frac{1}{N} \sum_{i=1}^{N} \left( \min\left( d( {e}_i,  {A}_{y_i}), d( {e}_i,  {B}_{y_i}) \right) \right)^2
\]

The inter-class loss is defined as a margin-based loss, encouraging the embeddings to maintain a margin $\delta$ from the line segments of incorrect classes:
\[
\text{Loss}_{\text{inter}} = \frac{1}{N} \sum_{i=1}^{N} \max(0, \delta - \min_{y_j \neq y_i} d( {e}_i, \overrightarrow{AB}_{y_j}))
\]

\subsection{Exponential Moving Average Update (EMA)}

We update the positions of the vertices $ {A}_k$ and $ {B}_k$ using an Exponential Moving Average (EMA) with a smoothing factor $\tau$:
\[
 {A}_k^{\text{new}} = \tau \cdot  {A}_k + (1 - \tau) \cdot  {A}_k^{\text{old}}
\]
\[
 {B}_k^{\text{new}} = \tau \cdot  {B}_k + (1 - \tau) \cdot  {B}_k^{\text{old}}
\]
where $\tau$ is a hyperparameter controlling the speed of vertex updates.

\subsection{Total Loss}
The DAAL loss $\zeta_{daal}$, which includes both intra-class and inter-class components is given by the equation:  
\begin{flushleft}
    \setlength{\fboxrule}{1pt} 
    \setlength{\fboxsep}{5pt} 
    \fbox{%
        \begin{minipage}{0.5\textwidth} 
            \vspace{10pt} 
            \begin{equation}
            \begin{split}
                \zeta_{daal} = & \frac{1}{N} \sum_{i=1}^{N} \left( \left( \min\left( d( {e}_i,  {A}_{y_i}), d( {e}_i,  {B}_{y_i}) \right) \right)^2 \right. \\
                & \left. + \lambda \max\left( 0, \delta - \min_{y_j \neq y_i} d( {e}_i, \overrightarrow{AB}_{y_j}) \right) \right)
            \end{split}
            \end{equation}
            \vspace{10pt} 
        \end{minipage}
    }
\end{flushleft}

The total loss is a weighted combination of the softmax classification loss $\zeta_{s}$ and The DAAL loss $\zeta_{daal}$ which is expressed as:
\begin{equation}
\zeta_{\text{total}} = \overbrace{\lambda_{s} \cdot \zeta_{s}}^{\text{Softmax Loss}} + \overbrace{\lambda_{\text{DAAL}} \cdot \zeta_{daal}}^{\text{DAAL Loss}}
\end{equation}

where $\lambda_s$ and $\lambda_{\text{DAAL}}$ are weighting factors controlling the contributions of the softmax and DAAL losses, respectively.

\subsection{Impact on Intra-Class Variance}

The adaptive line segment $\overrightarrow{AB}$ shrinks or extends based on the variance $\sigma_k$ of each class. For dense classes with low variance, the line segment contracts, ensuring tighter clusters in the embedding space. For classes with higher variance, the line segment extends, allowing for greater flexibility in embedding the spread-out samples.

Mathematically, the line segment adjusts as follows:

\begin{figure*}[htpb]
    \centering
    \setlength{\fboxrule}{0.5pt} 
    \setlength{\fboxsep}{5pt} 
    \fbox{\includegraphics[width=\textwidth]{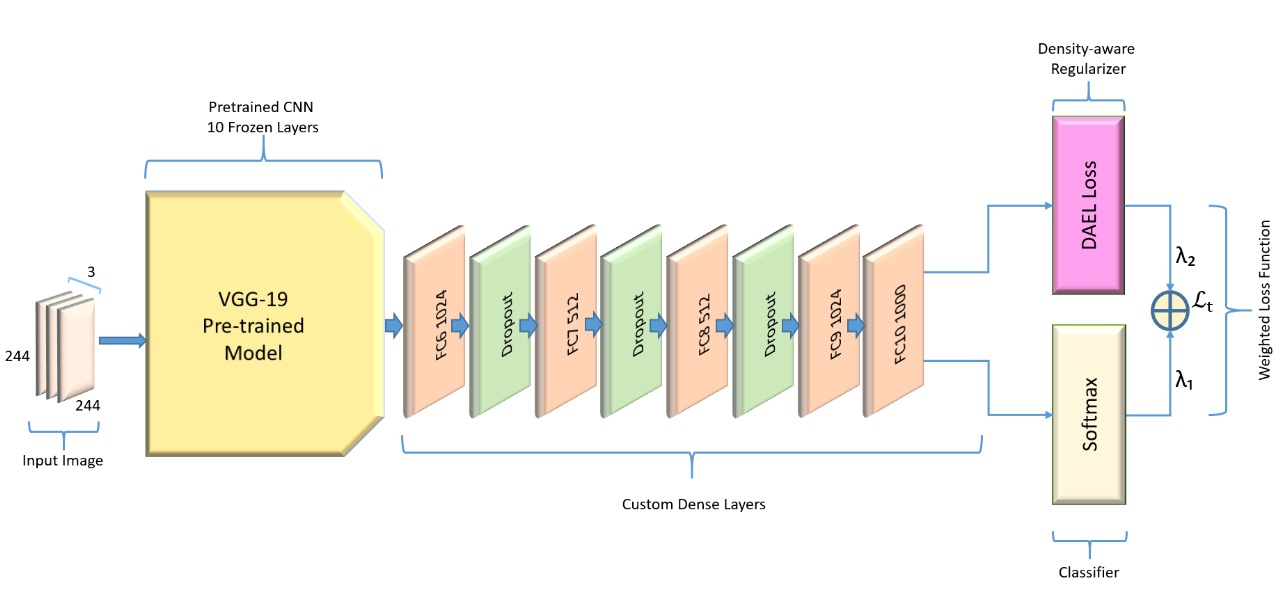}} 
    \caption{DAAL-DMA Model Architecture}
\end{figure*}

- If $\sigma_k$ is small, $ {A}_k$ and $ {B}_k$ are close to the centroid $ {c}_k$, leading to a tighter grouping of embeddings.
- If $\sigma_k$ is large, $ {A}_k$ and $ {B}_k$ move further from $ {c}_k$, promoting a more extended and flexible representation of the class in the embedding space.

This dynamic adjustment reduces the intra-class variance for dense classes while accommodating more spread-out classes, leading to improved performance in scenarios where class distribution varies significantly.

\section{Model Design}
\subsection{VGG19 Backbone Architecture}
We employed a pre-trained VGG19 model as the foundational architecture for our proposed design. VGG19 is widely recognized for its robust feature extraction capabilities, having been pre-trained on the large-scale ImageNet dataset. This model consists of 16 convolutional layers and three fully connected (dense) layers. The convolutional layers, as shown in fig 1, are organized into five distinct blocks, with each block followed by a max-pooling layer to progressively reduce the spatial dimensions while preserving relevant feature information.

The fully connected layers of the original VGG19, namely fc6, fc7, and fc8, are sized 4096, 4096, and 1000, respectively. In our modified architecture, we replace these dense layers to adapt to our dataset and task-specific needs. Importantly, the final layer in the original VGG19 applies a softmax activation for multi-class classification over 1000 categories. To retain the valuable feature representations learned by VGG19, we freeze the first ten convolutional layers during training, allowing only the later layers to be fine-tuned for our dataset. This approach ensures that our model leverages the rich, transferable features of VGG19 while adapting to the specific characteristics of our data.

\subsection{Classification and Density-Aware Loss Layers}
In our proposed model, as shown in fig 2, we append custom fully connected layers and a specialized loss layer to the VGG19 backbone. After flattening the output from the convolutional layers, we introduce multiple fully connected layers with 1024 and 512 units, each employing the Swish activation function. To prevent overfitting, we apply dropout regularization after each dense layer. The embeddings produced by these dense layers are then used both for classification and for calculating a novel loss function.

The classification head consists of a fully connected layer with a softmax activation function, producing logits that represent the predicted class probabilities. However, to enhance the discriminative power of the learned embeddings, we incorporate a Density-Aware Adaptive Line(DAAL) loss layer. This loss addresses the challenge of multi-modal feature distributions often present within a class. Many datasets exhibit intra-class variations, such as different poses or expressions in face recognition, which form multiple dense sub-clusters. The Density-Aware Adaptive Line Loss represents each class as an adaptive line segment, defined by two vertices: a centroid reflecting intra-class variance and a second point maintaining a margin from other classes.
\section{Hyperparameter Influence}

In the DAAL framework, key hyperparameters impact how the model balances intra-class variation and inter-class separation. Below is an explanation of how each hyperparameter influences the model, using the chosen values of margin \(\delta = 1.5\), EMA factor \(\tau = 0.001\), variance factor \(\eta = 5\), and loss weights \(\lambda_s = 1\) and \(\lambda_{daal} = 0.01\).

\subsection{DAAL Loss}

The DAAL loss is a sum of intra-class and inter-class loss terms. It is defined as:

\[
L_{\text{DAAL}} = \underbrace{\frac{1}{N} \sum_{i=1}^{N} d(\mathbf{e}_i, \overrightarrow{AB})^2}_{\text{intra-class loss}} + \underbrace{\frac{1}{N} \sum_{i=1}^{N} \max(0, \delta - d(\mathbf{e}_i, \overrightarrow{AB}))}_{\text{inter-class margin loss}}
\]

This loss penalizes embeddings that deviate too far from their class-defined line segment \(\overrightarrow{AB}\) (intra-class loss) and encourages separation between classes by enforcing a margin \(\delta\) (inter-class loss).

\subsection{Margin}

The margin \(\delta\) controls the separation between different class segments. With \(\delta = 1.5\), the model ensures sufficient inter-class separation, preventing embeddings from different classes from overlapping. This margin size provides enough separation without being overly restrictive, which is important when classes may share some common characteristics.

\subsection{EMA Factor}

The EMA factor \(\tau = 0.001\) controls how quickly the line segment vertices \(\overrightarrow{AB}\) adapt to changing class distributions. With a small value of \(\tau\), the updates to \(\mathbf{A}\) and \(\mathbf{B}\) are gradual, stabilizing the model over time. This ensures that the line segment can track class changes without overfitting to recent embeddings, thus accommodating diverse intra-class variations.

\subsection{Variance Factor}

The variance factor \(\eta = 5\) expands the line segments based on intra-class variance, which captures the spread of embeddings within each class. A higher \(\eta\) allows more flexibility in accommodating intra-class diversity, such as variations in lighting, pose, or expression in face recognition tasks. This factor enables the line segment to expand when there is a larger spread within a class, improving intra-class adaptability while maintaining structure.

\subsection{Lambda Factors}

The loss weights \(\lambda_s = 1\) and \(\lambda_{daal} = 0.01\) govern the balance between classification loss and DAAL loss. A higher weight for softmax loss (\(\lambda_s = 1\)) ensures that the primary focus remains on classification accuracy, while the smaller DAAL weight (\(\lambda_{daal} = 0.01\)) allows DAAL to shape the intra-class structure and improve separation without overwhelming the overall loss function. This balance ensures that the model maintains classification performance while still addressing intra-class variance and inter-class separation.

\subsection{Balancing Hyperparameters}

The chosen values for margin, variance factor, and EMA ensure that the model can handle multi-modal intra-class distributions while maintaining clear inter-class separation. The margin provides sufficient inter-class space, the variance factor allows intra-class flexibility, and the EMA factor stabilizes the line segments during training. The loss weightings ensure that the model focuses primarily on classification, while still benefiting from the structure imposed by DAAL loss.

The final loss function is:

\[
L_{\text{total}} = \lambda_s \cdot L_{\text{softmax}} + \lambda_{\text{daal}} \cdot L_{\text{DAAL}}
\]

where \(L_{\text{softmax}}\) is the softmax classification loss, and \(L_{\text{DAAL}}\) is the combined intra-class and inter-class DAAL loss.

\section{Experiments and Analysis}
\begin{table*}[ht]
    \centering
    \caption{PERFORMANCE COMPARISONS WITH THE STATE-OF-THE-ART METHODS IN TERMS OF NMI AND RECALL@K (\%) ON CARS196 AND CUB200 DATASET. . THE BEST PERFORMANCE ARE IN BOLD AND WE ALSO UNDERLINE THE PERFORMANCES OF THE BEST COMPETITORS. THE PERFORMANCE OF ALL THE MODELS WE DIRECTLY EXTRACT RESULTS REPORTED IN \cite{b20}}
     \label{tab:full_width_table_1}
    \begin{tabularx}{\textwidth}{|c|X|X|X|X|X|X|X|X|X|X|} 
        \hline
        \multicolumn{11}{|c|}{Cars196} \\ 
        \hline
        Paper & NMI & R@1 & R@2 & R@4 & R@8 & R@16 & R@32 & R@64 & R@128 & R@Average \\ 
        \hline 
        Tri \cite{b7}    & 47.23   & 42.54   & 53.94   & 65.74   & 75.06   & 82.40   & 88.70   & 93.17   & 96.42   & 74.74  \\
        LS \cite{b8}    & 56.88  & 52.98  & 65.70  & 76.01  & 84.27  & \-  & \-  & \-  & \-  & 69.74  \\
        NP \cite{b19}    & 57.29  & 56.52  & 68.42  & 78.01  & 85.70  & 91.19  & 94.81  & 97.38  & 98.83  & 83.85  \\
        Clu \cite{b74}    & 59.04  & 58.11  & 70.64  & 80.27  & 87.81  & \-  & \-  & \-  & \-  & 74.20  \\
        Con \cite{b59}    & 59.09  & 67.95  & 78.05  & 85.78  & 91.60  & 95.34  & 97.58  & 98.78  & 99.51  & 89.32  \\
        HDC \cite{b75}    & 62.71  & 71.42  & 81.85  & 88.54  & 93.40  & 96.59  & 98.16  & 99.21  & 99.67  & 91.10  \\
        DML-DA\textsubscript{tri}\cite{b20}    & 56.59  & 62.51  & 73.58  & 82.24  & 88.56  & 93.17  & 95.89  & 97.86  & 98.98  & 86.60  \\
        DML-DA\textsubscript{np} \cite{b20}     & 62.07  & 71.34  & 81.29  & 87.92  & 92.74  & 95.89  & 97.70  & 98.82  & 99.53  & 90.65  \\
        DML-DA\textsubscript{con} \cite{b20}     & \underline{65.17}  & \underline{77.62}  & \textbf{86.25}  & \textbf{91.71}  & \textbf{95.35}  & \textbf{97.54}  & \textbf{98.89}  & \textbf{99.37}  & \textbf{99.73}  & \textbf{93.30}  \\
        \hline
        \hline
        $DAEL-DML$   & \textbf{88.67}  & \textbf{80.96}  & 86.11  & 89.52  & 92.31  & 94.48  & 96.52  & 98.04  & 98.89  & 92.10 \\
        \hline
    \end{tabularx}
\end{table*}

\begin{table*}[ht]
    \label{tab:full_width_table_2}
    \centering
    \begin{tabularx}{\textwidth}{|c|X|X|X|X|X|X|X|X|X|X|} 
        \hline
        \multicolumn{11}{|c|}{CUB 200} \\ 
        \hline
        Paper & NMI & R@1 & R@2 & R@4 & R@8 & R@16 & R@32 & R@64 & R@128 & R@Average \\ 
        \hline 
        Tri \cite{b7}    & 50.99   & 39.57   & 51.74   & 63.35   & 74.14   & 82.98   & 89.53   & 94.80   & 97.65   & 74.22  \\
        LS \cite{b8}    & 56.50  & 43.57  & 56.55  & 68.59  & 79.63  & -  & -  & -  & -  & 62.08  \\
        NP \cite{b19}    & 57.41  & 47.30  & 59.57  & 70.75  & 80.98  & 88.28  & 93.50  & 96.79  & 98.43  & 79.45  \\
        Clu \cite{b74}    & 59.23  & 48.18  & 61.44  & 71.83  & 81.92  & -  & -  & -  & -  & 65.84  \\
        Con \cite{b59}    & 60.07  & 52.01  & 65.16  & 75.71  & 84.25  & 90.82  & 95.17  & 97.70  & 98.99  & 82.47  \\
        HDC \cite{b75}    & 60.78  & 52.50  & 65.25  & 76.01  & 85.03  & 91.10  & 95.34  & 97.67  & 99.09  & 82.74  \\
        DML-DA\textsubscript{tri} \cite{b20}    & 55.53  & 45.90  & 57.97  & 69.53  & 80.23  & 88.15  & 94.01  & 97.00  & 98.63  & 78.92  \\
        DML-DA\textsubscript{np} \cite{b20}     & 59.67  & 51.45  & 63.01  & 74.38  & 83.78  & 90.58  & 95.16  & 97.64  & 98.89  & 81.86  \\
        DML-DA\textsubscript{con} \cite{b20}     & \underline{62.32}  & \underline{55.64}  & \underline{66.96}  & \underline{77.92}  & \textbf{86.23}  & \textbf{92.10}  & \textbf{95.95}  & \textbf{98.11}  & \textbf{99.21}  & \underline{84.01}  \\
        \hline
        \hline
        $DAEL-DML$   & \textbf{81.78}  & \textbf{66.59}  & \textbf{74.36}  & \textbf{79.74}  & 84.81  & 89.08  & 93.05  & 95.77  & 97.82  & \textbf{85.15} \\
        \hline
    \end{tabularx}
\end{table*}
\subsection{Experimental Setup}
Our experiments were conducted on the Kaggle platform using two NVIDIA Tesla T4 GPUs with TensorFlow 2.16.1 leveraging its public datasets for model training and benchmarking. The Kaggle environment, pre-configured with necessary libraries like CUDA 12.3 and cuDNN 8, provided a seamless setup for running deep learning experiments, significantly reducing training time and ensuring reproducibility.
\subsection{Datasets}
We chose the CUB-2011 \cite{b71} and Cars196 \cite{b70} datasets for our experiments due to their established reputation as benchmark datasets in fine-grained visual categorization.

\textbf{Cars196}  The Cars196 dataset is a benchmark for fine-grained visual recognition, comprising 16,185 images across 196 classes of car models. It includes 8,144 training and 8,041 testing images, all resized to 224x224 pixels for deep learning. With notable variations in color, viewpoint, and background, the dataset challenges accurate classification, necessitating robust feature extraction methods for distinguishing similar models while maintaining clear inter-class separability.

\textbf{CUB-200-211} The CUB-200-2011 dataset, a benchmark for fine-grained image classification, consists of 11,788 bird images across 200 species, showcasing significant intra-class variability in color, pose, and background. It includes 5,994 training and 5,794 testing images, all resized to 224x224 pixels for deep learning compatibility. This dataset is crucial for evaluating advanced deep learning models in distinguishing closely related classes, making it invaluable for fine-grained visual recognition research.




\subsection{Evaluation Metrics and Compared Methods}

\textbf{Evaluation Metrics.} For performance evaluation, we utilized the Normalized Mutual Information (NMI) metric \cite{b72} to assess the alignment between the predicted clusters and the true labels. NMI is derived as the ratio of mutual information to the average entropy of both the clusters and the labels, serving as a reliable indicator of clustering quality. Additionally, we employed Recall@K (R@K) \cite{b73} to quantitatively assess our model's performance in the k-nearest neighbor retrieval task. The Recall@K score is set to 1 if an image of the same class is found among the k-nearest neighbors for a given test image query; otherwise, it is 0. The overall R@K metric is calculated by averaging the Recall@K scores across all image queries in the test set. To provide a more comprehensive evaluation, we also incorporated Recall@Average (R@Average), which aggregates Recall@K scores for various values of K, offering insights into the model's retrieval capabilities at different precision levels. All metrics, including NMI and Recall@K, are computed using the code provided in [?].

\begin{figure*}[htbp] 
    \centering
    \setlength{\fboxrule}{1pt} 
    \setlength{\fboxsep}{5pt} 
    \fbox{\includegraphics[width=1.0\textwidth]{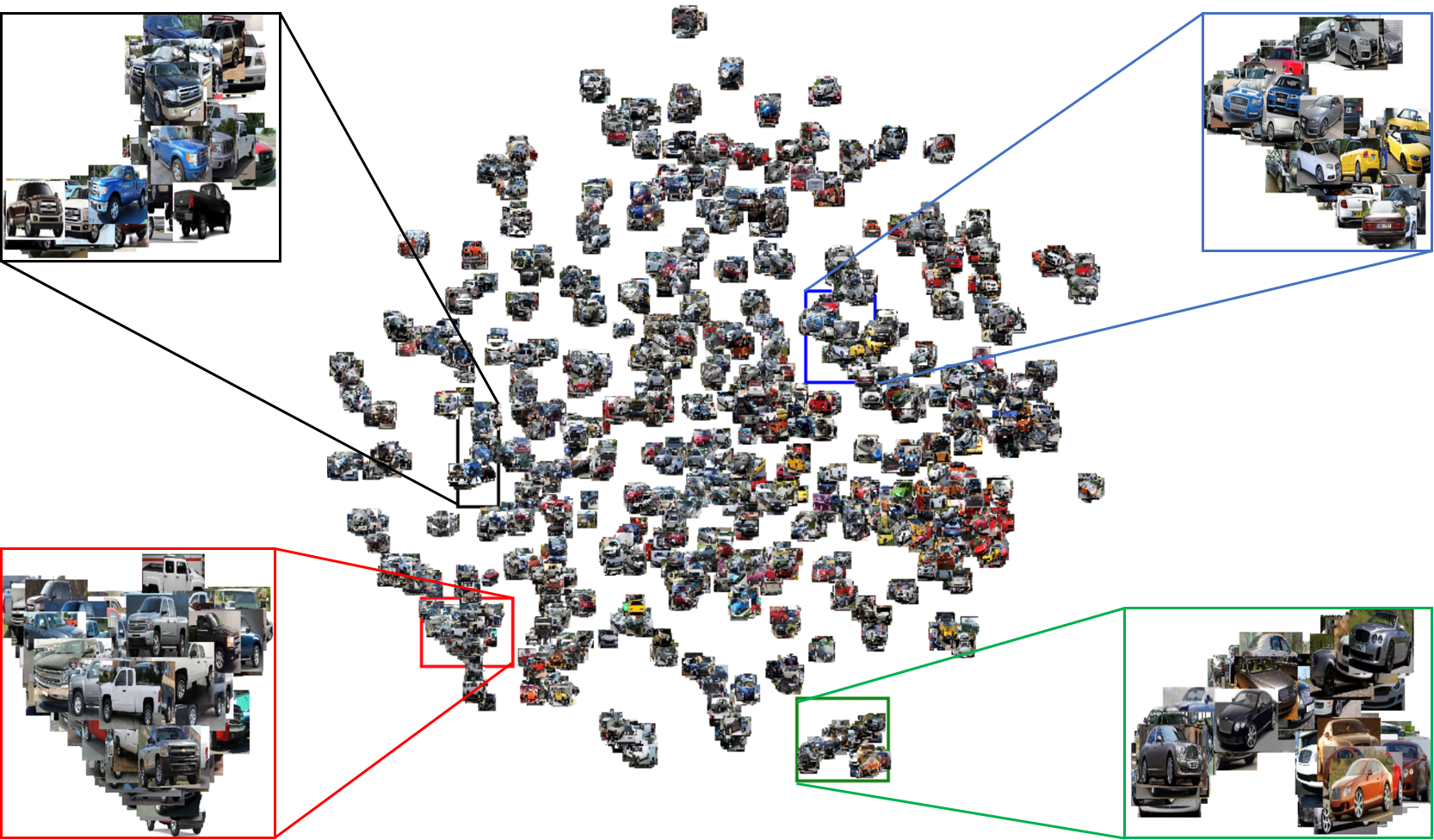}} 
    \caption{Barnes-Hut t-SNE visualization \cite{b50} illustrates the image embeddings learned by our DAAL-DML model on the test set of the Cars196 dataset. By incorporating density adaptivity into the DML training process, DAAL-DML achieves an effective balance between inter-class similarity and intra-class diversity, which strengthens the model's generalization ability. Despite the cars in the same class having different colors, poses, and backgrounds, the model successfully preserves this variation across different distributions within each cluster, as shown in the zoomed images, while maintaining clear inter-class separation.}
\end{figure*}

\textbf{Compared methods}
We have evaluated our model against the following methods: Triplet \cite{b7}, Lifted Struct \cite{b8}, N-Pair \cite{b19}, Clustering \cite{b74}, Contrastive \cite{b59}, HDC \cite{b75}, DML-DA\textsubscript{con} \cite{b20}, DML-DA\textsubscript{np} \cite{b20} and DML-DA\textsubscript{tri} \cite{b20}. The results are documented in Table 1, with the top section presenting the results for Cars196 and the bottom section for CUB-200-2011.

\subsection{Performance comparison}
Table 1 presents the performance comparisons of various state-of-the-art methods in terms of NMI and Recall@K\% on the \textbf{CARS196} dataset. The \textbf{DAEL-DML} model achieves an impressive NMI score of \textbf{88.67}, outperforming all other methods by a significant margin of \textbf{23.50\%} over the previous best, which is \textbf{DML-DA}$_{\textit{con}}$ at \textbf{65.1}7. In terms of Recall@1, \textbf{DAEL-DML} also leads with a value of \textbf{80.96}, exceeding \textbf{DML-DA}$_{\textit{con}}$ \textbf{77.62} by \textbf{3.34\%}. While \textbf{DAEL-DML} performs robustly in most Recall@K metrics, it trails slightly behind the top competitors in Recall@2, Recall@4, and Recall@8, Recall@16, Recall@32, Recall@64, Recall@128, indicating competitive performance in these categories.

On the \textbf{CUB200} dataset (Table 2), the \textbf{DAEL-DML} model again showcases superior performance with an NMI of \textbf{81.78}, the best among all listed models and exceeding the previous best by \textbf{19.46\%}. Notably, we excel in Recall@1, achieving a score of \textbf{66.59}, surpassing \textbf{DML-DA}$_{\textit{con}}$ at \textbf{55.64} by \textbf{10.95\%}. The performance in Recall@2 and Recall@4 also shows a lead, with \textbf{DAEL-DML} scoring \textbf{74.36} and \textbf{79.74}, respectively, compared to \textbf{DML-DA}$_{\textit{con}}$ \textbf{66.96} and \textbf{77.92}, yielding differences of \textbf{7.40\%} and \textbf{1.82\%}. Importantly, \textbf{DAEL-DML} outperforms competitors in Recall@Average, attaining a score of \textbf{85.15}, which is \textbf{1.14\%} higher than the best competitor.

Overall, the \textbf{DAEL-DML} model demonstrates significant advancements in both datasets, particularly excelling in NMI and maintaining a competitive edge across Recall@K metrics. It consistently surpasses leading methods in Recall@Average, particularly on the \textbf{CUB200} dataset, affirming its robustness and reliability in deep metric learning tasks.

\begin{figure*}[htbp] 
    \centering
    \setlength{\fboxrule}{1pt} 
    \setlength{\fboxsep}{10pt} 
    \fbox{\includegraphics[width=1.0\textwidth]{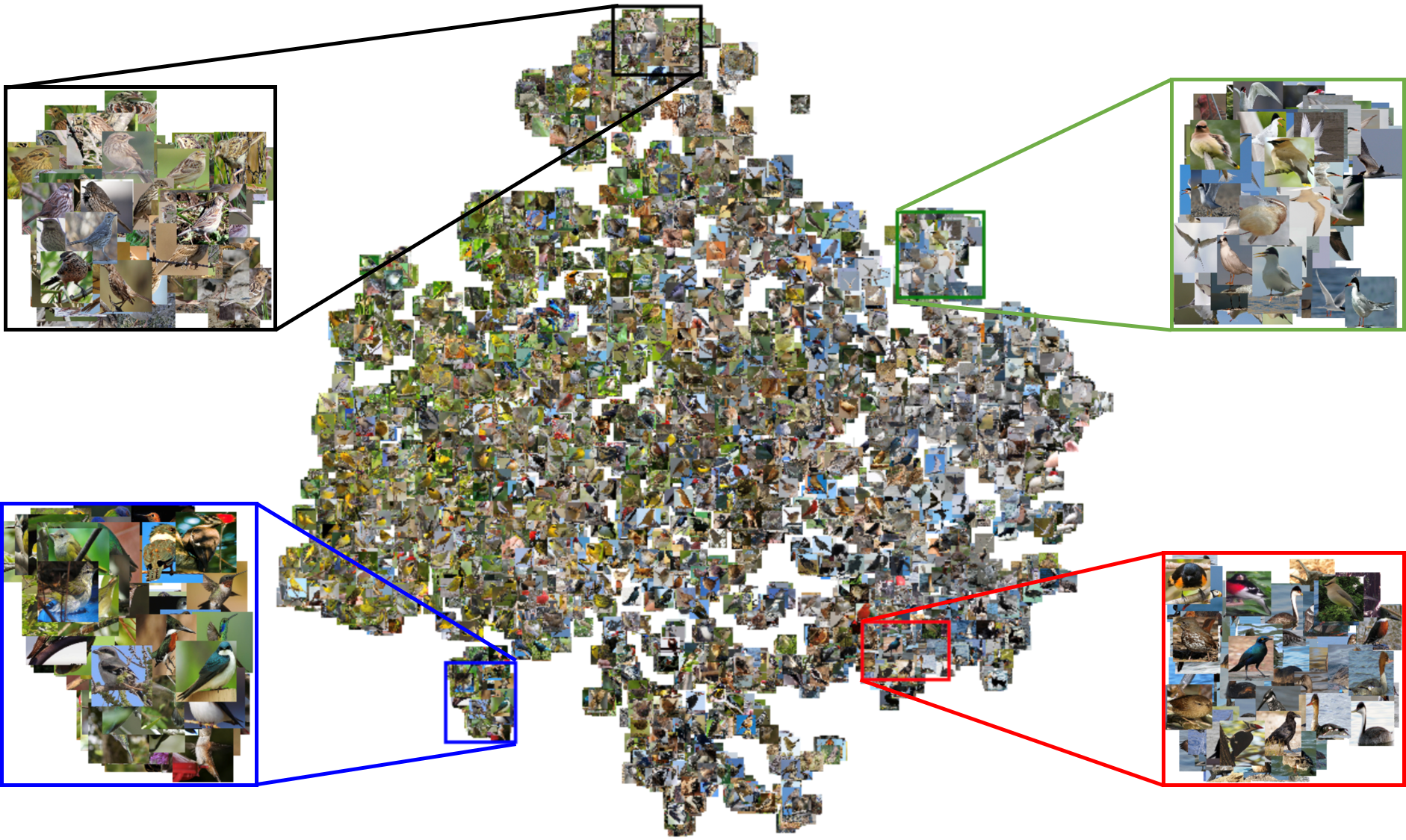}} 
    \caption{Barnes-Hut t-SNE visualization \cite{b50} illustrates the image embeddings learned by our DAAL-DML model on the test set of the CUB-200-2011 dataset. By incorporating density adaptivity into the DML training process, DAAL-DML achieves an effective balance between inter-class similarity and intra-class diversity, which strengthens the model's generalization ability. Despite the birds in the same class having different colors, poses, and backgrounds, the model successfully preserves this variation across different distributions within each cluster, as shown in the zoomed images, while maintaining clear inter-class separation.}
\end{figure*}

\subsection*{Embedding representation visualization}
The effectiveness of our DAAL-DML model is illustrated through Barnes-Hut t-SNE visualizations \cite{b50} of image embeddings learned from both the CUB-200-2011 and Cars196 datasets. In the case of the CUB-200-2011 dataset, the t-SNE visualization reveals that the model adeptly captures the intricate relationships among bird species, clustering semantically similar birds together despite substantial variations in viewpoint and background. By incorporating density adaptivity into the deep metric learning (DML) training process, DAAL-DML achieves a harmonious balance between inter-class similarity and intra-class variation, significantly enhancing the model's generalization capability. This adaptability ensures that the distribution of embeddings within each class is preserved, allowing for a more representative clustering of the various attributes present within the species.

Similarly, the t-SNE representation for the Cars196 dataset shows that cars within the same class—characterized by a range of colors, poses, and backgrounds—are effectively grouped. The model maintains this distribution across different variations within each cluster, ensuring clear inter-class separation while still respecting the inherent diversity of the data. These visualizations confirm that the learned embeddings are not only discriminative but also robust to the variability found in real-world data, highlighting the DAAL loss's effectiveness in preserving class distributions during the embedding process.

\section{Discussions and future work}
In future work, we aim to enhance the model's generalization capability by implementing an elastic line as a density-aware intra-class variation regularizer. This approach has the potential to improve performance by allowing for a more nuanced representation of class distributions. Additionally, integrating this regularizer with margin-based softmax could yield even better results, as it would leverage the strengths of both techniques.

Furthermore, we plan to introduce polylines instead of straight lines for mapping into the embedding space. This shift could effectively capture the natural relationships between intra-class variations, providing a more flexible representation that adapts to the complexities of real-world data. By accommodating the inherent diversity within each class, the model will be better equipped to manage variations before and after the mapping process, ultimately leading to improved embedding performance.
\\
\\
\textbf{Funding}
This research received no external funding.
\\
\textbf{Data Availability}
The data supporting this study are available on GitHub at \href{https://github.com/HadushHailu/DAAL-DML}{GitHub Repository}

\section*{Declarations}
\textbf{Conflict of interest} The authors declare that they have no conflict of interest.

\textbf{Open Access} This article is licensed under a Creative Commons Attribution 4.0 International License, which permits use, sharing,adaptation, distribution and reproduction in any medium or format, as long as you give appropriate credit to the original author(s) and the source, provide a link to the Creative Commons licence, and indicate if changes were made. The images or other third party material in this article are included in the article’s Creative Commons licence, unless indicated otherwise in a credit line to the material. If material is not included in the article’s Creative Commons licence and your intended use is not permitted by statutory regulation or exceeds the permitted use, you will need to obtain permission directly from the copyright holder. To view a copy of this licence, visit http://creativecommons.org/licenses/by/4.0/.

\vspace{12pt}


\begin{thebibliography}{00}
\bibitem{b1} R. Manmatha, Chao-Yuan Wu, Alexander J. Smola, and
Philipp Kr¨ahenb¨uhl. Sampling matters in deep embedding
learning. In ICCV, pages 2859–2867, 2017

\bibitem{b2} R. Salakhutdinov and G. Hinton. Learning a nonlinear em-
bedding by preserving class neighbourhood structure. In
AISTATS, 2007.
\bibitem{b3} . R. Hershey, Z. Chen, J. L. Roux, and S. Watanabe, “Deep clustering:
Discriminative embeddings for segmentation and separation,” in IEEE
International Conference on Acoustics, Speech and Signal Processing.
IEEE, 2016, pp. 31–35.
\bibitem{b4} L. Ma, X. Yang, and D. Tao, “Person re-identification over camera
networks using multi-task distance metric learning,” IEEE Transactions
on Image Processing, vol. 23, no. 8, pp. 3656–3670, 2014.
\bibitem{b5} F. Schroff, D. Kalenichenko, and J. Philbin. FaceNet: A
unified embedding for face recognition and clustering. In
CVPR, 2015.
\bibitem{b6} Omkar M. Parkhi, Andrea Vedaldi, and Andrew Zisserman.
Deep face recognition. In BMVC, pages 41.1–41.12, 2015. 
\bibitem{b7} Kilian Q. Weinberger and Lawrence K. Saul. Distance met-
ric learning for large margin nearest neighbor classification.
Journal of Machine Learning Research, 10:207–244, 2009.
\bibitem{b8} Hyun Oh Song, Yu Xiang, Stefanie Jegelka, and Silvio
Savarese. Deep metric learning via lifted structured feature
embedding. In CVPR, pages 4004–4012, 2016.
\bibitem{b9} G. Chechik, V. Sharma, U. Shalit, and S. Bengio. Large scale
online learning of image similarity through ranking. Journal
of Machine Learning Research, 11:1109–1135, 2010.
\bibitem{b10} S. Bell and K. Bala. Learning visual similarity for prod-
uct design with convolutional neural networks. ACM Trans.
Graph., 34(4):98:1–98:10, 2015.
\bibitem{b11} Oren Rippel, Manohar Paluri, Piotr Doll´ar, and Lubomir D.
Bourdev. Metric learning with adaptive density discrimina-
tion. ICLR, 2016.
\bibitem{b12}  Qi Qian, Rong Jin, Jinfeng Yi, Lijun Zhang, and Shenghuo
Zhu. Efficient distance metric learning by adaptive sampling
and mini-batch stochastic gradient descent (SGD). Machine
Learning, 99(3):353–372, 2015. 
\bibitem{b13} Z. Li and J. Tang, “Weakly supervised deep metric learning for
community-contributed image retrieval,” IEEE Transactions on Multi-
media, vol. 17, no. 11, pp. 1989–1999, 2015.
\bibitem{b14} Irjanto NS, Surantha N (2020) Home security system with face
recognition based on convolutional neural network. Int J Adv
Comput Sci Appl. https://doi.org/10.14569/IJACSA.2020.
\bibitem{b15} Seal A, Bhattacharjee D, Nasipuri M, et al (2013a) Thermal
human face recognition based on gappypca. In: 2013 IEEE 2nd
international conference on image information processing (ICIIP-
2013), IEEE, p 597–600
\bibitem{b16} Sohn K (2016) Improved deep metric learning with multi-class
n-pair loss objective. In: Advances in neural information pro-
cessing systems, vol 29
\bibitem{b17} J. Weston, S. Bengio, and N. Usunier. WSABIE: scaling up
to large vocabulary image annotation. In IJCAI, pages 2764–
2770, 2011.

\bibitem{b18} Z. Zhang and V. Saligrama, “Zero-shot learning via joint latent similarity
embedding,” in Proceedings of the IEEE Conference on Computer Vision
and Pattern Recognition, 2016, pp. 6034–6042.
\bibitem{b19} K. Sohn, “Improved deep metric learning with multi-class n-pair loss
objective,” in Advances in Neural Information Processing Systems, 2016,
pp. 1857–1865.

\bibitem{b20} Y. Li, T. Yao, Y. Pan, H. Chao and T. Mei, "Deep Metric Learning With Density Adaptivity," in IEEE Transactions on Multimedia, vol. 22, no. 5, pp. 1285-1297, May 2020, doi: 10.1109/TMM.2019.2939711
\bibitem{b21} Eric P. Xing, Andrew Y. Ng, Michael I. Jordan, and Stuart J.
Russell. Distance metric learning with application to cluster-
ing with side-information. In NIPS, pages 505–512, 2002.
\bibitem{b22} Chopra S, Hadsell R, LeCun Y (2005) Learning a similarity
metric discriminatively, with application to face verification. In:
2005 IEEE computer society conference on computer vision and
pattern recognition (CVPR’05), IEEE, p 539–546
\bibitem{b23} X. He, Y. Zhou, Z. Zhou, S. Bai and X. Bai, "Triplet-Center Loss for Multi-view 3D Object Retrieval," 2018 IEEE/CVF Conference on Computer Vision and Pattern Recognition, Salt Lake City, UT, USA, 2018, pp. 1945-1954, doi: 10.1109/CVPR.2018.00208.
\bibitem{b24} X. Zhang, F. Zhou, Y. Lin, and S. Zhang. Embedding label
structures for fine-grained feature representation. In CVPR,
2016.
\bibitem{b25} S. Chopra, R. Hadsell, and Y. LeCun. Learning a similarity
metric discriminatively, with application to face verification.
In CVPR, 2005.
\bibitem{b26} Khalifa A, Abdelrahman AA, Strazdas D et al (2022) Face
recognition and tracking framework for human-robot interaction.
Appl Sci 12(11):5568
\bibitem{b27} Johnson M, Bradshaw JM (2021) How interdependence explains
the world of teamwork. A systems engineering approach to
realizing synergistic capabilities, engineering artificially intelli-
gent systems. Springer, Cham, pp 122–146
\bibitem{b28} Seal A, Ganguly S, Bhattacharjee D, et al (2013b) Thermal
human face recognition based on haar wavelet transform and
series matching technique. In: Multimedia processing, commu-
nication and computing applications: proceedings of the 1st
international conference, ICMCCA, Springer, p 155–167
\bibitem{b29} R. Hadsell, S. Chopra, and Y. LeCun, “Dimensionality reduction by
learning an invariant mapping,” in Proceedings of the IEEE Conference
on Computer Vision and Pattern Recognition, vol. 2. IEEE, 2006, pp.
1735–1742.
\bibitem{b30} H. Kim, T. Mei, H. Byun, and T. Yao, “Exploiting web images for
video highlight detection with triplet deep ranking,” IEEE Transactions
on Multimedia, vol. 20, no. 9, pp. 2415–2426, 2018.
\bibitem{b31} Qi Qian, Rong Jin, Shenghuo Zhu, and Yuanqing Lin. Fine-
grained visual categorization via multi-stage metric learning.
In CVPR, pages 3716–3724, 2015. 
\bibitem{b32} Deng J, Guo J, Xue N, et al (2019) Arcface: additive angular
margin loss for deep face recognition. In: Proceedings of the
IEEE/CVF conference on computer vision and pattern recogni-
tion, p 4690–4699
\bibitem{b33} M. H. Kiapour, X. Han, S. Lazebnik, A. C. Berg, and T. L.
Berg. Where to buy it: Matching street clothing photos in
online shops. In ICCV, 2015.
\bibitem{b34} Y. Li, H. Su, C. R. Qi, N. Fish, D. Cohen-Or, and L. J.
Guibas. Joint embeddings of shapes and images via CNN im-
age purification. ACM Trans. Graph., 34(6):234:1–234:12,
2015.
\bibitem{b35} Huang Y, Wang Y, Tai Y, et al (2020) Curricularface: adaptive
curriculum learning loss for deep face recognition. In: proceed-
ings of the IEEE/CVF conference on computer vision and pattern
recognition, p 5901–5910
\bibitem{b36}  Jiao J, Liu W, Mo Y et al (2021) Dyn-arcface: dynamic additive
angular margin loss for deep face recognition. Multimed Tools
Appl 80(17):25741–25756
\bibitem{b37} Boutros F, Damer N, Kirchbuchner F, et al (2022) Elasticface:
elastic margin loss for deep face recognition. In: Proceedings of
the IEEE/CVF conference on computer vision and pattern
recognition, p 1578–1587
\bibitem{b38} J. Wang, Y. Song, T. Leung, C. Rosenberg, J. Wang,
J. Philbin, B. Chen, and Y. Wu. Learning fine-grained im-
age similarity with deep ranking. In CVPR, 2014.
\bibitem{b39} O. Russakovsky, J. Deng, H. Su, J. Krause, S. Satheesh,
S. Ma, Z. Huang, A. Karpathy, A. Khosla, M. S. Bernstein,
A. C. Berg, and F. Li. ImageNet large scale visual recog-
nition challenge. International Journal of Computer Vision,
115(3):211–252, 2015.
\bibitem{b40} Wen Y, Zhang K, Li Z, et al (2016) A discriminative feature
learning approach for deep face recognition. In: European con-
ference on computer vision, Springer, p 499–515
\bibitem{b41} Liu W, Wen Y, Yu Z, et al (2017) Sphereface: deep hypersphere
embedding for face recognition. In: Proceedings of the IEEE
conference on computer vision and pattern recognition,
p 212–220
\bibitem{b42} Y. Taigman, M. Yang, M. Ranzato, and L. Wolf. DeepFace:
Closing the gap to human-level performance in face verifica-
tion. In CVPR, 2014.
\bibitem{b43} Wang H, Wang Y, Zhou Z, et al (2018) Cosface: large margin
cosine loss for deep face recognition. In: Proceedings of the IEEE
conference on computer vision and pattern recognition,
p 5265–5274
\bibitem{b44} Q. Qian et al., "SoftTriple Loss: Deep Metric Learning Without Triplet Sampling," 2019 IEEE/CVF International Conference on Computer Vision (ICCV), Seoul, Korea (South), 2019, pp. 6449-6457, doi: 10.1109/ICCV.2019.00655.
\bibitem{b45} Wen Y, Liu W, Weller A, et al (2022) Sphereface2: binary
classification is all you need for deep face recognition. In:
International conference on learning representations
\bibitem{b46} Guillaumin M, Verbeek J, Schmid C (2009) Is that you? metric
learning approaches for face identification. In: IEEE 12th inter-
national conference on computer vision, IEEE, p 498–505
\bibitem{b47} Y. Sun, Y. Chen, X. Wang, and X. Tang, “Deep learning face rep-
resentation by joint identification-verification,” in Advances in Neural
Information Processing Systems, 2014, pp. 1988–1996.
\bibitem{b48} KAYA M, BİLGE HŞ. Deep Metric Learning: A Survey. Symmetry. 2019; 11(9):1066. https://doi.org/10.3390/sym11091066 
\bibitem{b49} Y. Li, Y. Pan, T. Yao, H. Chao, Y. Rui, and T. Mei, “Learning
click-based deep structure-preserving embeddings with visual attention,”
ACM Transactions on Multimedia Computing, Communications, and
Applications, vol. 15, no. 3, p. 78, 2019.
\bibitem{b50} O. Rippel, M. Paluri, P. Dollar, and L. Bourdev, “Metric learning
with adaptive density discrimination,” in International Conference on
Learning Representations, 2016
\bibitem{b51} arkhi OM, Vedaldi A, Zisserman A (2015) Deep face
recognition
\bibitem{b52} Weifeng Ge, Weilin Huang, Dengke Dong, and Matthew R.
Scott. Deep metric learning with hierarchical triplet loss. In
ECCV, pages 272–288, 2018.
\bibitem{b53} Y. Cui, F. Zhou, Y. Lin, and S. J. Belongie. Fine-grained
categorization and dataset bootstrapping using deep metric
learning with humans in the loop. In CVPR, 2016.
\bibitem{b54} T. Yao, T. Mei, and C.-W. Ngo, “Learning query and image similarities
with ranking canonical correlation analysis,” in Proceedings of the IEEE
International Conference on Computer Vision, 2015, pp. 28–36.
\bibitem{b55} E. Ustinova and V. Lempitsky, “Learning deep embeddings with his-
togram loss,” in Advances in Neural Information Processing Systems,
2016, pp. 4170–4178.
\bibitem{b56} C. Huang, C. C. Loy, and X. Tang, “Local similarity-aware deep feature
embedding,” in Advances in Neural Information Processing Systems,
2016, pp. 1262–1270.
\bibitem{b57} Y. Bai, Y. Lou, F. Gao, S. Wang, Y. Wu, and L. Duan, “Group sensitive
triplet embedding for vehicle re-identification,” IEEE Transactions on
Multimedia, vol. 20, no. 9, pp. 2385–2399, 2018.

\bibitem{b58} Kavalionak H, Gennaro C, Amato G et al (2019) Distributed
video surveillance using smart cameras. J Grid Comput 17:59–77
\bibitem{b59} S. Bell and K. Bala, “Learning visual similarity for product design with
convolutional neural networks,” ACM Transactions on Graphics, vol. 34,
no. 4, p. 98, 2015.
\bibitem{b60}  Liu W, Wen Y, Yu Z, et al (2016) Large-margin softmax loss for
convolutional neural networks. In: Proceedings of the 33rd 
international conference on international conference on machine learning-vol 48, p 507–516
\bibitem{b61} Liu H, Zhu X, Lei Z, et al (2019b) Adaptiveface: adaptive margin
and sampling for face recognition. In: Proceedings of the IEEE/
CVF conference on computer vision and pattern recognition,
p 11947–11956
\bibitem{b62} Deng J, Zhou Y, Zafeiriou S (2017) Marginal loss for deep face
recognition. In: Proceedings of the IEEE conference on computer
vision and pattern recognition workshops, p 60–68
\bibitem{b63} P. Cui, S. Liu, and W. Zhu, “General knowledge embedded image
representation learning,” IEEE Transactions on Multimedia, vol. 20,
no. 1, pp. 198–207, 2018.
\bibitem{b64} Seal A, Bhattacharjee D, Nasipuri M, et al (2012) Minutiae from
bit-plane sliced thermal images for human face recognition. In:
Proceedings of the international conference on soft computing for
problem solving (SocProS 2011), Springer, p 113–124
\bibitem{b65} Chopra, R. Hadsell, and Y. LeCun. Learning a similarity
metric discriminatively, with application to face verification.
In CVPR, volume 1, pages 539–546, 2005.
\bibitem{b66} T. Yao, T. Mei, and Y. Rui, “Highlight detection with pairwise deep
ranking for first-person video summarization,” in Proceedings of the
IEEE Conference on Computer Vision and Pattern Recognition, 2016.
\bibitem{b67} Oinar C, Le BM, Woo SS (2022) Kappaface: adaptive additive
angular margin loss for deep face recognition. arXiv preprint
arXiv:2201.07394
\bibitem{b68} He L, Wang Z, Li Y, et al (2020) Softmax dissection: towards
understanding intra-and inter-class objective for embedding
learning. In: Proceedings of the AAAI conference on artificial
intelligence, p 10957–10964
\bibitem{b69} Kim M, Jain AK, Liu X (2022) Adaface: quality adaptive margin
for face recognition. In: Proceedings of the IEEE/CVF confer-
ence on computer vision and pattern recognition, p 18750–18759

\bibitem{b70} Jonathan Krause, Michael Stark, Jia Deng, and Li Fei-Fei.
3d object representations for fine-grained categorization. In
4th International IEEE Workshop on 3D Representation and
Recognition (3dRR-13), Sydney, Australia, 2013

\bibitem{b71} C. Wah, S. Branson, P. Welinder, P. Perona, and S. Belongie.
The Caltech-UCSD Birds-200-2011 Dataset. Technical report, 2011.

\bibitem{b72} C. D. Manning, P. Raghavan, H. Schutze ¨ et al.
, Introduction to information retrieval. Cambridge university press, 2010, vol. 16, no. 1.

\bibitem{b73} H. Jegou, M. Douze, and C. Schmid. Product quantization for nearest neighbor search. In PAMI, 2011.


\bibitem{b74} H. Oh Song, S. Jegelka, V. Rathod, and K. Murphy, “Deep metric
learning via facility location,” in Proceedings of the IEEE Conference
on Computer Vision and Pattern Recognition, 2017, pp. 5382–5390.


\bibitem{b75} Y. Yuan, K. Yang, and C. Zhang, “Hard-aware deeply cascaded embedding,” in Proceedings of the IEEE International Conference on
Computer Vision, 2017, pp. 814–823.

\end{thebibliography}
\end{document}